\documentclass{article}
\usepackage{fullpage}

\usepackage{hyperref}

\usepackage{amsmath}
\usepackage{amsfonts}
\usepackage{amsbsy}
\usepackage{graphicx} % more modern
\usepackage{subfigure} 

\usepackage{bm}
\usepackage[usenames,dvipsnames]{color}
\usepackage{mdframed}
\usepackage{accents}

\newcommand{\ubar}[1]{\ensuremath{\underline{#1}}}
\newcommand{\ssty}{\ensuremath{\scriptstyle}}  % Make math small
\newcommand{\sssty}{\ensuremath{\scriptscriptstyle}}  % Make math smaller

\newcommand{\dorder}[1]{\ensuremath{{\sssty (#1)}}}

\newcommand{\limarrow}[2]{\ensuremath{\xrightarrow[#1\to#2]{}}}

\newcommand{\R}{\mathbb{R}}

\newcommand{\spt}[1]{\mathcal{#1}}
\newcommand{\mat}[1]{{\boldsymbol{#1}}}

\newcommand{\e}{\mat{e}}

\newcommand{\g}{\mat{g}}

\newcommand{\h}{\mat{h}}

\newcommand{\q}{\mat{q}}
\newcommand{\qd}{{\dot{\mat{q}}}}
\newcommand{\qdd}{{\ddot{\mat{q}}}}
\newcommand{\x}{\mat{x}}

\newcommand{\xd}{\dot{\mat{x}}}

\newcommand{\z}{\mat{z}}
\newcommand{\zd}{\dot{\mat{z}}}
\newcommand{\zdd}{\ddot{\mat{z}}}

\newcommand{\ed}{\dot{\e}}

\newcommand{\A}{\mat{A}}
\newcommand{\B}{\mat{B}}
\newcommand{\C}{\mat{C}}
\newcommand{\D}{\mat{D}}

\newcommand{\I}{\mat{I}}
\newcommand{\J}{\mat{J}}
\newcommand{\Jp}{\mat{J}_{\phi}}

\newcommand{\M}{\mat{M}}

\newcommand{\U}{\mat{U}}

\newcommand{\zero}{\mat{0}}

\newcommand{\mH}{\mat{H}}

\newcommand{\wt}[1]{\widetilde{#1}}

\newtheorem{theorem}{Theorem}

\newtheorem{corollary}{Corollary}

\title{\LARGE \bf
On the Fundamental Importance of Gauss-Newton in Motion Optimization
}

\author{
  Nathan Ratliff\thanks{Lula Robotics Inc., Seattle, WA, USA} \and
  Marc Toussaint\thanks{Machine Learning and Robotics Lab, University of Stuttgart, Germany} \and
  Jeannette Bohg\thanks{Autonomous Motion Department, Max Planck Institute for Intelligent Systems, T\"{u}bingen, Germany} \and 
  Stefan Schaal\footnotemark[3]
}

\begin{document}

\maketitle
\thispagestyle{empty}
\pagestyle{empty}

%%%%%%%%%%%%%%%%%%%%%%%%%%%%%%%%%%%%%%%%%%%%%%%%%%%%%%%%%%%%%%%%%%%%%%%%%%%%%%%%
\begin{abstract}

% Intro
Hessian information speeds convergence substantially in motion optimization.
The better the Hessian approximation the better the convergence.  But how good
is a given approximation theoretically? How much are we losing?  This paper
addresses that question and proves that for a particularly popular and
empirically strong approximation known as the Gauss-Newton approximation, we
actually lose very little---for a large class of highly expressive objective
terms, the true Hessian actually \textit{limits to} the Gauss-Newton Hessian
quickly as the trajectory's time discretization becomes small.  This result
both motivates it's use and offers insight into computationally efficient
design.  For instance, traditional representations of kinetic energy exploit
the generalized inertia matrix whose derivatives are usually difficult to
compute.  We introduce here a novel reformulation of rigid body kinetic energy
designed explicitly for fast and accurate curvature calculation.  Our theorem
proves that the Gauss-Newton Hessian under this formulation efficiently
captures the kinetic energy curvature, but requires only as much computation as
a single evaluation of the traditional representation.  Additionally, we
introduce a technique that exploits these ideas implicitly using Cholesky
decompositions for some cases when similar objective terms reformulations exist
but may be difficult to find.  Our experiments validate these findings and
demonstrate their use on a real-world motion optimization system for high-dof
motion generation.

\end{abstract}
%%%%%%%%%%%%%%%%%%%%%%%%%%%%%%%%%%%%%%%%%%%%%%%%%%%%%%%%%%%%%%%%%%%%%%%%%%%%%%%%

\section{Introduction}
\label{sec:Introduction}

Hessian information is playing an increasingly central role in trajectory
optimization for motion generation, especially in high degree-of-freedom
systems with non-trivial dynamics and environmental constraints.  But
generally, Hessian calculations can be hard, especially when kinematic maps or
inertia matrices are involved.  Successful motion optimization methods have
gone to lengths to avoid large portions of them.  Many, such as CHOMP
\cite{RatliffCHOMP2009}, STOMP \cite{Kalakrishnan_RAIIC_2011}, and TrajOpt
\cite{AbbeelTrajopt2013}, largely ignore the curvature induced by kinematic
maps and focus solely on the connectivity encoded in the interactions between
neighboring configurations within smoothness terms of discrete-time trajectory
representations.  Others, such as iLQG
\cite{iLQGTodorov05,DRCIntegratedSystemTodorov2013}, KOMO
\cite{KOMOToussaint2014}, and RieMO \cite{RIEMORatliff2015ICRA}, have found
empirically that Gauss-Newton approximations tend to be sufficient for fast
convergence.  Unfortunately, there's little theory backing the various
approximations in the context of motion optimization; we're often left to rely
on empirical justification and intuition.  We don't really know how much the
approximation hinders performance.

This paper takes a deeper look at this theoretical problem.  Specifically, we
analyze how the Gauss-Newton approximation relates to the true Hessian for a
widely used and expressive class of objective terms represented as functions of
task space time derivatives.\footnote{Here, we define a task space as the
co-domain of any differentiable map from the configuration space. See
Section~\ref{sec:TheoreticalBackground}.} We show that for this class of
functions, when numerically integrated across the trajectory, the true Hessian
\textit{converges to} the Gauss-Newton approximation as the time-discretization
becomes increasingly fine.  This Gauss-Newton Hessian\footnote{ Throughout this
  paper, we use the term Gauss-Newton \textit{Hessian} rather than Gauss-Newton
\textit{approximation} to emphasize its fundamental contribution to the
curvature representation for these problems.  } is actually a fundamental
representation of problem curvature.  Importantly, this class of objective
terms includes discrete approximations to the geodesic energy through a task
space, and can therefore theoretically represent any sort of Riemannian
geometric model describing fundamental properties of the robot such Euclidean
end-effector motion, kinetic energy (see
Section~\ref{sec:ReformulatingKineticEnergy}), and the geometry of the
workspace \cite{RIEMORatliff2015ICRA}.

Our proof also shows that this limit converges at a rate of $O(\Delta t^{2k})$,
where $k$ is the minimal order of the task space time-derivatives used in the
terms. This Gauss-Newton Hessian is known to capture the entirety of the
first-order Riemannian Geometry of these types of terms
\cite{RIEMORatliff2015ICRA}; here, we show that the Gauss-Newton Hessian
actually \textit{is} the Hessian in the limit of finer discretization.

We also show that some previously computationally difficult objective terms can
be reformulated in light of these observations to exploit this efficient
curvature calculation.  As an example, we reformulate rigid body kinetic energy
of the robot as a squared velocity norm through a higher-dimensional Euclidean
task space to leverage this result.  The calculation of the Gauss-Newton
Hessian under this formulation requires only Jacobians of kinematic maps, and
is only as computationally complex as calculating the inertia matrix $\M(\q)$
required for a \textit{single evaluation} of the traditional formulation
$\spt{K}(\q, \qd) = \frac{1}{2}\qd^T\M(\q)\qd$.

Moreover, since any Riemannian metric can be represented as a Pullback metric
of a mapping into a higher-dimensional Euclidean space (see
Section~\ref{sec:TheoreticalBackground}), we can approximately and 
implicitly exploit these
results, even when we don't have access to the full task map, using
Cholesky decompositions of the Riemannian metric.

Our experiments provide an empirical verification of these theoretical
results, as well as a full system validation on a real-world 8
degree-of-freedom manipulation platform where we use motion optimization with
Gauss-Newton Hessians to solve problems such as obstacle avoidance and reaching
to grasp preshapes that conform to the overall shape
of the object (formulated as a single optimization).

\section{Motion Optimization with Finite-Differencing}
\label{sec:TheoreticalBackground}

This section reviews the general motion optimization formulation studied
in this paper as well as some basic concepts from 
Riemannian geometry useful for the discussion. For details, we point to
\cite{RIEMORatliff2015ICRA}.

The discrete-time motion optimization problem can be generically represented
as a constrained optimization problem defined over $k^{th}$-order
Markov cliques \cite{KOMOToussaint2014}. 
We denote the full discrete trajectory\footnote{For effective modeling 
  of finite-differenced derivatives, we often use a fixed 
  prefix configuration $\q_0$ and an extra (non-fixed) suffix configuration $\q_{T+1}$,
  while still letting the terminal potential act on $\q_T$.} 
as $\xi = (\q_1,\ldots,\q_T)$ with $\q_t\in\R^d$ and 
time index $t$ in integer units of
the the discretization $\Delta t > 0$. Each $k^{\textrm{th}}$-order 
clique 
is a sequence of $k+1$ configurations starting at index $\tau_t$
(with $\tau_{t+1} = \tau_t + 1$), denoted\footnote{Semicolons, here and elsewhere, 
denote Matlab-style vector stacking. Note that we use $\tau_t$ to denote
the start index for the clique window at time $t$ because it's useful
for the clique to be centered at the right time step 
when $\Delta t$ and time are important to the model.
} 
$\q_t^c = (\q_{\tau_t};\q_{\tau_t+1};\ldots;\q_{\tau_t+k})$.
In full, the optimization problem takes the form
\begin{align}
  \min_{\xi} \sum_{t=1}^T \spt{L}_t(\q_t^c)
  \ \ \
  \textrm{s.t.}\ \ 
  \left\{
  \begin{array}{l}
    \g_t(\q_t^c) \leq \zero \\\nonumber
    \h_t(\q_t^c) = \zero
  \end{array}
  \right.
  \forall t.
\end{align}
We use Augmented Lagrangian methods for constrained optimization which, in an
inner loop, add the violated constraints into the objective as penalty
functions\footnote{The penalty functions use Lagrange multiplier estimates to
  effectively shift the zero point of the penalty away from the constraint
boundary opposite the direction of constraint violation to pull the minimizer
onto the constraint surface without requiring infinitely large penalty scalars
\cite{NocedalWright2006}.} to form an unconstrained proxy objective which is
subsequently optimized using Newton's method. 

Each time step's objective term $\spt{L}_t(\q_t^c)$ usually consists of a
collection of modeling costs $\wt{l}(\q_t^c)$, many of which take the form
$\wt{l}(\q_t^c) = l(\phi(\q_t^c))$, where $\phi$ is a differentiable nonlinear
mapping of the entire clique $\q_t^c$ to some other space.  Often these clique
maps are constructed from \textit{task maps} of the form
$\phi:\R^d\rightarrow\R^m$; we use $\phi$ to denote both the map and its
configuration-wise application to the entire clique $\z_t^c = \phi(\q_t^c)$.
These task maps may be mappings from the configuration space to points and axes
on the robot's body, or from the configuration space to the distance between a
pair of fingers.  We even treat the identity map as a task map for uniformity,
making the configuration space, itself, a task space.  When $\z = \phi(\q)$
represents a task map, we call the map's co-domain the \textit{task space}; the
corresponding mapped variable $\z\in\R^m$ is said to reside in the that task
space.

Further, each of these task space cost functions $l(\phi(\q_t^c)) = l(\z_t^c)$
are often functions of time derivatives calculated as finite-differences on the
clique. We denote $k^{\textrm{th}}$-order finite-differencing matrices
(operating on a task space clique) by $\z_t^\dorder{k} = \D^\dorder{k}\z_t^c$.
For instance, if the clique is defined over triples of configurations $\z_t^c =
(\z_{t-1}; \z_t; \z_{t+1})$, the first- and second-order finite-differencing
matrices take the form $\D^\dorder{1} = \frac{1}{\Delta t}[-\I,\:\I,\:\zero]$
and $\D^\dorder{2} = \frac{1}{\Delta t^2}[\I,\:-2\I,\:\I]$ so that
multiplication with $\z_t^c$ calculates the velocity and acceleration,
respectively.  Using this notation, we can denote functions defined on task
space derivatives as $l(\z_t^c) =
f(\D^\dorder{k_1}\z_t^c,\ldots,\D^\dorder{k_l}\z_t^c)$.

{
\renewcommand{\Jp}{\ensuremath{\J_{\z_t^c}}}
The \textit{full Hessian} of a general task space clique function of 
the form $l(\phi(\q_t^c)) = l(\z_t^c)$ is
\begin{align*}
  \nabla^2_{\q_t^c} l(\phi({\q_t^c})) =
    \Jp^T\Big(\nabla^2 l(\z_t^c)\Big)\Jp
    + \left[\frac{\partial \Jp}{\partial \q_t^c}\right]
      \nabla l(\z_t^c),
\end{align*}
where in this context $\J_{\z_t^c}$ denotes the Jacobian of the full clique map
$\z_t^c = \phi(\q_t^c)$.  The derivative of the Jacobian
$\frac{\partial\Jp}{\partial\q}$ is a third-order tensor and is, therefore,
computationally expensive---it's equivalent to calculating $m$ separate
$n\times n$ Hessians. The \textit{Gauss-Newton Hessian} simply removes the term
that require this third-order tensor.
}

There's a strong connection between task maps and Riemannian Geometry.  For our
purposes, the Riemannian Geometry of a space is represented by a symmetric
positive definite matrix $\A(\q)$ that varies smoothly with $\q$. It defines a
norm on the tangent space\footnote{More generally, an inner product, but we're
primarily concerned with norms here. The tangent space is effectively the space
of velocities here.} so that $\|\qd\|_\A^2 = \qd^T\A(\q)\qd$. Importantly,
given any task map $\z = \phi(\q)$, since $\zd = \Jp\qd$, we can rewrite
tangent space norms in the task space under metric $\B(\z)$ as tangent space
norms in the configuration space $\|\zd\|_\B^2 = \|\Jp\qd\|_\B^2 =
\qd^T\left(\Jp^T\B\Jp\right)\qd$ with Riemannian metric $\A(\q) = \Jp^T\B\Jp$.
This derived metric is called the Pullback metric
\cite{LeeSmoothManifolds2012}. 

The right choice of task map $\phi$ can be a very expressive model of the
problem's geometry.  And, typically, terms defined on derivatives through a
task space, satisfy the preconditions of Theorem~\ref{thm:MainTheorem},
allowing us to leverage Gauss-Newton Hessians to efficiently capture and
leverage the curvature of the problem.  Moreover, the Nash-Embedding theorem
\cite{NashEmbeddingTheorem1956} tells us that the geometry of every
non-Euclidean Riemannian manifold can be modeled as a mapping to a
higher-dimensional Euclidean task space.  Such a task map can be hard to find,
a situation addressed in Section~\ref{sec:CholeskyApproximations}, but, as
Section~\ref{sec:ReformulatingKineticEnergy} shows, finding such an equivalent
task map, especially given Theorem~\ref{thm:MainTheorem}, can greatly reduce
the computational burden of representing and exploiting problem curvature.

\section{The Convergence of Hessians to Gauss-Newton}
\label{sec:TheoreticalConvergence}

This section states and proves the primary theoretical results of this paper.
The theorem is stated in a fairly general form, which makes the proof somewhat
more difficult to follow.  For a simpler theoretical statement covering a
useful special case with a more verbose and explicit proof, see
\cite{RIEMORatliff2015}.  The proof presented here may be skipped on first
reading without loss of continuity.

In what follows, we denote the full Hessian of a function $\spt{F}(\xi)$
defined on the entire trajectory as $\nabla^2\spt{F}(\xi)$ and the
corresponding Gauss-Newton Hessian as $\nabla_{\mathrm{GN}}\spt{F}(\xi)$.  See
Section~\ref{sec:TheoreticalBackground} for additional notational definitions.

\subsection{Basic theorem statements and corollaries}

The main theorem states that discrete approximations of trajectory integrals
with objective terms defined on finite-differenced
time derivatives of the trajectory mapped into some task space
have Hessians that converge quickly to the Gauss-Newton Hessian 
as the time-discretization becomes small.

\begin{mdframed}
\begin{theorem} \label{thm:MainTheorem}
  Consider
  a sequence of objective terms of the form
  \begin{align}
    \spt{F}(\xi) = \sum_{t=1}^T f(\D^\dorder{k_1}\z_t^c,\:\ldots,\:\D^\dorder{k_l}\z_t^c)\:\Delta t
  \end{align}
  defined over task-space cliques, where $f(\cdot)$ is independent of
  $\Delta t$. Then $\nabla^2\spt{F}(\xi)\rightarrow\nabla^2_{\mathrm{GN}}\spt{F}(\xi)$
  as $\Delta t\rightarrow 0$ at a rate of $O(\Delta t^{2k})$ where $k = \min_i \{k_i\}_{i=1}^l$.
\end{theorem}
\end{mdframed}
The proof of this theorem is given in Section~\ref{sec:TheoremProof}. A direct corollary of 
this result is that squared task space velocity and acceleration norms---common
modeling tools---have Hessians that converge to the Gauss-Newton approximation
at well defined rates that increase with the order of the derivative.
\begin{mdframed}
\begin{corollary} \label{cor:Corollary}
  Let $\zd_t = \D^\dorder{1}\z_t^c$ and $\zdd_t = \D^\dorder{2}\z_t^c$. Then objectives of the 
  form
  \begin{align*}
    \spt{F}_{\zd}(\xi) = \sum_{t=1}^T \frac{1}{2}\|\zd_t\|^2\:\Delta t
    \ \ \textrm{and}\ \
    \spt{F}_{\zdd}(\xi) = \sum_{t=1}^T \frac{1}{2}\|\zdd_t\|^2\:\Delta t
  \end{align*}
  have Hessians that converge to the Gauss-Newton approximation at the rates of 
  $O(\Delta t^2)$ and $O(\Delta t^4)$, respectively, independent of the particular
  choice of differentiable map defining $\z_t = \phi(\q_t)$.
\end{corollary}
\end{mdframed}

\subsection{Proof of Theorem \ref{thm:MainTheorem}}
\label{sec:TheoremProof}

Due to space restrictions, we prove only convergence of the block diagonal
entries for the special case where each clique function is defined on only
single derivative $f_t(\z_t^\dorder{k})$. It's straightforward to show that the
off-diagonal Hessian blocks are always equivalent to the Gauss-Newton
approximation, independent of $\Delta t$, and the proof of mixed
time-derivatives is similar.  In this section, we denote the dimensionality of
the task space as $m > 0$ with $\phi(\q) = \z \in\R^m$ denoting the task map.

We first note that finite-differencing matrices for $k^{\textrm{th}}$ order
derivatives take the blockwise form
\begin{align*}
  \D^\dorder{k} = \left[ \D_0^\dorder{k}\ \D_1^\dorder{k}\ \cdots\ \D_k^\dorder{k}\right]
  \ \ \textrm{with}\ \
  \D_i^\dorder{k} = \frac{\sigma_i^\dorder{k}}{\Delta t^k} \I,
\end{align*}
where $\sigma_i^\dorder{k}\in\R$ are constants and $\I\in\R^{m\times m}$.
Consider any sequence of clique terms of the form
\begin{align}
  \spt{F}(\xi) = \sum_{t=1}^T l(\z_t^c) = \sum_{t=1}^T f(\D^\dorder{k}\z_t^c).
\end{align}
In the following, we use subscripts to indicate which clique 
a function is applied to. 
For instance, we denote $l(\z_t^c)$ as $l_t$.

Let $\spt{N}_t = \{\tau\:|\:\z_t \in\z_\tau^c\}$
be the set of all indices of cliques containing the task space variable $\z_t$
and let $\ubar{t} = \min\spt{N}_t$ be the smallest
of those indices. 
Generally, in terms of $l$ (defined 
directly on the clique variables), 
the Hessian block corresponding to the single configuration $\q_t$ is
\begin{align} \label{eqn:GenericHessian}
  \nabla^2_{\q_t}\spt{F}
  = \J_t^T\left(\sum_{i=0}^k \nabla^2_{\z_t} l_{\ubar{t}+i}\right)\J_t
  + \left[\frac{\partial\J_t}{\partial\q_t}\right]\sum_{i=0}^k \nabla_{\z_t} l_{\ubar{t}+i}.
\end{align}
Note that $\frac{\partial \J_t}{\partial \q_t} = \frac{\partial^2 \phi_t}{\partial \q_t}$
is a third-order tensor, so the product is a tensor product.
The first term in Equation~\ref{eqn:GenericHessian} is already the Gauss-Newton
Hessian, so what remains to be shown is that the second term gets vanishingly
small relative to the first as $\Delta t$ gets small.

Now we can expand these expressions in terms of $f$ (defined
on the finite-differenced time derivatives).
Denoting the derivative of $f$ with respect to the time-derivative 
variable as $\z^\dorder{k} = \D^\dorder{k}\z^c$ as $\nabla_{\z^\dorder{k}} f$,
the sum of first derivatives in Equation~\ref{eqn:GenericHessian} becomes
\begin{align*}
  \sum_{i=0}^k \nabla_{\z_t} l_{\ubar{t}+i}
  &= \sum_{i=0}^k {\D_{k-i}^\dorder{k}}^T \nabla_{\z^\dorder{k}} f(\D^\dorder{k}\z_{\ubar{t}+i}^{c})
  = \ubar{\D}^\dorder{k}\g_t^c,
\end{align*}
where the $i^{\textrm{th}}$ $m$-dimensional segment of this new clique variable 
$\g_t^c$ is $\nabla_{\z^\dorder{k}} f(\D^\dorder{k}\z_{\ubar{t}+i}^{c})$,
and $\ubar{\D}^\dorder{k} = \left[\D_k^\dorder{k}\ \D_{k-1}^\dorder{k}\ \ldots\ \D_0^\dorder{k}\right]$ 
evaluates the $k^{\textrm{th}}$-order finite-differencing time derivative
approximation for time running in reverse (note that each $\D_i^\dorder{k}$ is symmetric). 
This sum, therefore, limits to the following kinematic quantity
\begin{align} \label{eqn:GradientLimit}
  \ubar{\D}^\dorder{k}\g_t^c
  \ \ \ \limarrow{\Delta t}{0}\ \ \ 
    -\frac{d^k}{dt^k} \left(
      \frac{\partial f}{\partial \z^\dorder{k}}\left(\z^\dorder{k}(t)\right)
      \right),
\end{align}
where $\z^\dorder{k}(t) = \frac{d^k}{dt^k}\phi(\q(t))$.

{
\newcommand{\alphaik}{\ensuremath{\alpha_i^{\scriptscriptstyle (k)}}}
The sum over second derivatives in Equation~\ref{eqn:GenericHessian}
is
\begin{align*}
  \sum_{i=0}^k \nabla^2_{\z_t}l(\z_{\ubar{t}+i}^\dorder{k})
  &= \sum_{i=0}^k {\D_{k-i}^\dorder{k}}^T\left(\nabla^2_{\z^\dorder{k}}f\big(\D^\dorder{k}\z_{\ubar{t}+i}^c\big)\right) \D_{k-i}^\dorder{k}
  \\ \nonumber
  &= \frac{S_k}{\Delta t^{2k}}
      \sum_{i=0}^k 
        \alphaik\left(\nabla^2_{\z^\dorder{k}}f\big(\D^\dorder{k}\z_{\ubar{t}+i}^c\big)\right),
\end{align*}
where $S_k = \sum_i\left(\sigma_{k-i}^\dorder{k}\right)^2$ and 
$\alphaik = \left(\sigma_{k-i}^\dorder{k}\right)^2/S_k$ are both constant
with respect to $\Delta t$, and $\alphaik$ form normalized nonnegative
weights with $\alphaik \geq 0$ and 
$\sum_i\alphaik = 1$.
Each of the Hessians $\C_{ti}^\dorder{k} = \nabla^2_{\z^\dorder{k}}f(\D^\dorder{k}\z_{\ubar{t}+i}^c)$ limit to the same 
value $\C_t^\dorder{k} = \nabla_{\z^\dorder{k}}^2f(\z^\dorder{k}(t))$ as $\Delta t\rightarrow 0$, 
so the weighted average of Hessians,
itself, limits to $\C_t^\dorder{k}$.
Since this expression we're evaluating---the
first term of Equation~\ref{eqn:GenericHessian}---is 
also the Gauss-Newton Hessian,
and we've just shown that it scales with $O(\frac{1}{\Delta t^{2k}})$ (getting 
large as $\Delta t$ gets small), we need to scale the expressions by $O(\Delta t^{2k})$
to understand their limiting behavior so 
that the Gauss-Newton portion limits to a finite nonzero value.
The scaled expression in Equation~\ref{eqn:GenericHessian}, using the derived 
calculations, becomes
\begin{align} \nonumber
  \frac{\Delta t^{2k}}{S_k}\nabla^2_{\q_t}\spt{F}
  &= \J_t^T\left(\sum_{i=0}^k\alphaik\C_{ti}^\dorder{k}\right)\J_t
    + \frac{\Delta t^{2k}}{S_k}\left[\frac{\partial\J_t}{\partial\q_t}\right]\ubar{\D}^\dorder{k}\g_t^c.
\end{align}
The first term is always the (scaled) Gauss-Newton Hessian and 
limits to $\J_t^T\C_t^\dorder{k}\J_t$, while the second
term approaches zero at a rate of $O(\Delta t^{2k})$ since the unscaled 
portion of that term approaches the kinematic limit 
in Equation~\ref{eqn:GradientLimit}.
}
\hfill$\square$

\section{Reformulating Kinetic Energy for Efficient Curvature Calculations}
\label{sec:ReformulatingKineticEnergy}

Kinetic energy is an important quantity in motion, 
but interestingly it's not usually modeled in kinematic optimization strategies 
for motion planning such as CHOMP \cite{RatliffCHOMP2009}, 
STOMP \cite{Kalakrishnan_RAIIC_2011}, TrajOpt \cite{AbbeelTrajopt2013},
or AICO \cite{ToussaintTrajOptICML2009}. 
We can largely attribute that discrepancy to its computational complexity,
especially with respect to Hessian calculation.
The kinetic energy is most commonly expressed in the configuration space as
\begin{align} \label{eqn:TraditionalKineticEnergy}
  \spt{K}(\q, \qd) = \frac{1}{2}\qd^T\M(\q) \qd,
\end{align}
where $\M(\q)$ is the symmetric positive definite inertia matrix.  Generally,
taking derivatives of $\M(\q)$ is hard.  There are some clever algorithms for
calculating such derivatives \cite{ClosedFormDynamicsDerivativesOtt2013}, but
fundamentally, the first derivative will always be a third order tensor, and
the second derivative will always be a fourth order tensor.  Computing these in
practice is both difficult and computationally expensive.

But we can interpret $\M(\q)$ as a positive definite Riemannian metric, and
Equation~\ref{eqn:TraditionalKineticEnergy} as a generalized squared velocity
norm.  The Nash Embedding Theorem (see Section~\ref{sec:TheoreticalBackground})
suggests that there exist a nonlinear map to some higher dimensional task space
under which Euclidean squared velocity norms correctly reflect this metric
weighted velocity. And importantly, if we find such a map,
Corollary~\ref{cor:Corollary} say the Hessian of these terms when summed across
the trajectory converges to the Gauss-Newton Hessian at a rate of $O(\Delta
t^2)$.

One such map is straightforward, but computationally intractable. 
If we could enumerate all 
particles across the entire robot's 
body $\spt{R} = \{\x_i\}_{i=1}^N$, the kinetic energy of the system is simply
\begin{align}
  \spt{K}(\q, \qd) = \sum_{i=1}^N \frac{1}{2} m_i \|\xd_i\|^2.
\end{align}
This expression is a squared velocity norm through a task space of the form
\begin{align}
  \phi_\spt{K}(\q) = \left[
    \begin{array}{c}
      \sqrt{m_1}\:\x_1(\q)  \\
      \sqrt{m_2}\:\x_2(\q)  \\
      \vdots \\
      \sqrt{m_N}\:\x_N(\q)
    \end{array}
  \right].
\end{align}
Unfortunately, $N$ is technically on the order of $10^{23}$ (Avogadro's number---this
sum is usually best approximated as an integral)
and even discrete approximations would require a large number of ``proxy'' particles
to make a good representation.

However, we can simplify this expression by 
leveraging the same rigid body structure here that's typically 
used to derive the rigid body inertia matrix. 
Suppose $\rho(\mat{u})$ with $\mat{u} = (u_1,u_2,u_3)$ 
describes the mass density of a rigid body in 
coordinates of some orthonormal basis $\{\e_1, \e_2, \e_3\}$ relative 
to the center-of-mass $\x_c$. (Here we're taking $\x_c,\e_1, \e_2, \e_3$ to be
described in some inertial world frame of reference, but fixed
relative to the rigid body so that the coordinates $\mat{u} = (u_1,u_2,u_3)$ 
always represent
a local description.\footnote{$\e_i$ constitute the columns of a
rotation matrix from local to world coordiantes; their Jacobians can 
be easily calculated via cross products with joint axes.}) Any given point on the robot's body $\spt{R}$
can be represented in world coordinates as 
$\x = \x_c + u_1\e_1 + u_2\e_2 + u_3\e_3 \in\spt{R}$.
In these coordinates, as Appendix~\ref{sec:ExploitingRigidBodyStructure} 
shows in more detail, the kinetic energy integral over the rigid body
decomposes as 
\begin{align} \nonumber
  \frac{1}{2}\int_\spt{R}\|\xd\|^2 \rho\:d\x
  &= \frac{1}{2}\int \rho(\mat{u})\|\xd_c + u_1\ed_1 + u_2\ed_2 + u_3\ed_3\|^2 d\mat{u}
  \\\label{eqn:RigidBodyEnergyDecomposition}
  &= \frac{1}{2}M\|\xd_c\|^2 + \sum_{ij} \frac{1}{2}b_{ij} \ed_i^T\ed_j.
\end{align}
where $M = \int_\spt{R} \rho\:d\x$ is the total mass, and the quantities 
$b_{ij}$ are given by 
\begin{align}
  b_{ij} = \int u_iu_j\:\rho(\mat{u})\:d\mat{u}.
\end{align}
These quantities $b_{ij}$ form entries in a matrix $\B$ that we call the
\textit{Distributional Inertia Matrix}.  This matrix describes the mass
\textit{distribution} of the body rather than the moments of inertia (note its
structural similarity to the covariance matrix of a probability distribution).
This Distributional Inertia Matrix is just a linear transformation away from
the more traditional form as described in
Appendix~\ref{sec:RelationshipBetweenInertiaMatrices}, and the principle
directions that diagonalize the matrix are the same. If the axes of
representation are chosen as these principle directions $\e_1^*,\e_2^*,\e_3^*$,
then all off-diagonal entries of $\B$ vanish, and the above expression reduces
to
\begin{align}
  &\spt{K}(\q, \qd) =
  \frac{1}{2}M\|\xd_c\|^2
  +
  \frac{1}{2}\sum_{i=1}^3 b_{i} \|\ed_i^*\|^2,
\end{align}
where $b_i = b_{ii} = \int u_i^2\:\rho(\mat{u})\:d\mat{u}$ denotes the moment 
of inertia of the axis $\e_i$ as though it were a infinitesimally thin rod
of material rotating around the center-of-mass.
Note that the axes $\e_i^*$ are normalized 
and the velocities $\ed_i^*$ always reflect only rotational components 
which are always orthogonal to the axis, itself. The norm $\|\ed_i^*\|$
is, therefore, the angular velocity of the rotation of that axis.

Thus, under this representation, we can construct a task map of the form
\begin{align}
  \phi_{\spt{K}}(\q) = \left[
    \begin{array}{c}
      \sqrt{M}\:\x_c \\
      \sqrt{b_1}\:\e_1^* \\
      \sqrt{b_2}\:\e_2^* \\
      \sqrt{b_3}\:\e_3^*
    \end{array}
  \right],
\end{align}
built entirely from kinematic maps of the rigid body.  Denoting this
\textit{Rigid Body Inertial Map} as $\z_{\spt{K}} = \phi_\spt{K}(\q)$, the
kinetic energy reduces to just $\frac{1}{2}\|\zd_{\spt{K}}\|^2$. This
formulation represents the kinetic energy of a rigid body as the squared
velocity norm through a 12 dimensional space.  Its curvature, therefore,
primarily comes from its Gauss-Newton Hessian, which, in this case, is computed
entirely from kinematic Jacobians.  Note that under the original formulation,
the inertia matrix, itself, already consists of a similar sum of kinematic
Jacobians;\footnote{Note that Gauss-Newton Hessians are essentially Pullback
metrics \cite{RIEMORatliff2015ICRA}.} calculating even its gradient, therefore,
requires second-order derivatives of these kinematic functions, and calculating
its Hessian is even harder.  On the other hand, computing the Gauss-Newton
Hessian of this reformulated kinetic energy is the same order of complexity as
just \textit{one evaluation} of the inertia matrix $\M(\q)$.

Notice that since the axes are orthogonal to one another, we can 
implement this representation using $\e_3^* = \e_1^* \times \e_2^*$.
This formulation takes 9 numbers to describe and is, therefore,
a minimal representation of rigid body kinetic energy.\footnote{
  The traditional representation requires the linear center-of-mass
  velocity, the angular momentum, and the rotation of the 
  inertia matrix into the world frame. Each of those quantities requires
  3 parameters in a minimal representation totaling 9 parameters in all.
}

\section{Cholesky Approximations for Arbitrary Metrics}
\label{sec:CholeskyApproximations}

Sometimes velocities through task space are 
most easily represented as metric scaled velocities through the configuration space
of the form 
$l(\q, \qd) = \frac{1}{2}\qd^T\A(\q)\qd$ for some smooth 
symmetric positive definite Riemannian metric $\A(\q)$.
The Nash Embedding theorem suggests
that all Riemannian metrics can be represented as the Pullback some Euclidean
task space metric (see Section~\ref{sec:TheoreticalBackground},
but sometimes it's difficult to find such a task map.
This section shows that the true Hessian of these terms 
can be approximated even without access to this unknown task map,
using only Cholesky factorizations of the metric $\A(\q)$.

Suppose $\z = \phi(\q)$ is a differentiable task map. The pullback metric
is $\A(\q) = \Jp^T\Jp$. For this argument, we assume the map is sufficient
to produce a fully positive definite metric. 
We first note that since the Cholesky Decomposition produces
$\A = \C^T\C$, the upper triangular matrix $\C$ is just an orthogonal
coordinate transformation away from the task map's Jacobian:
$\C^T\C = \Jp^T\Jp \Rightarrow (\Jp\C^{-1})^T(\Jp\C^{-1}) = \I$, which means
$\U = \Jp\C^{-1}$ has mutually orthogonal columns and $\Jp = \U\C$.

For simplicity, we describe this result specifically for terms 
of the form 
\begin{align} \label{eqn:SimpleTaskVelocity}
  l(\q_{t-1}, \q_t) = \frac{1}{2}\|\phi(\q_t) - \phi(\q_{t-1})\|^2,
\end{align}
where we assume we don't have access to $\phi$, but we can evaluate the 
Pullback metric $\A(\q) = \Jp^T\Jp$ directly as a function of $\q$.
The gradient takes the form
\begin{align*}
  \nabla l_t =
  \left[
  \begin{array}{c}
    \scriptstyle -\J_{\phi_{t-1}}^T(\phi(\q_t) - \phi(\q_{t-1})) \\
    \scriptstyle \J_{\phi_t}^T(\phi(\q_t) - \phi(\q_{t-1}))
  \end{array}
  \right]
  \approx
  \left[
    \begin{array}{c}
      \scriptstyle -\A_{t-1}(\q_t - \q_{t-1}) \\
      \scriptstyle \A_t(\q_t - \q_{t-1})
    \end{array}
  \right],
\end{align*}
where the approximation on the right comes from linearizing both $\phi(\q_{t-1})$ and 
$\phi(\q_{t})$ around $\q_{t-1}$ for the first entry and around $\q_t$ for the second entry.
Moreover, the Gauss-Newton Hessian is
\begin{align} \nonumber
  \nabla^2_{\textrm{GN}} l_t
  &=
  \left[
    \begin{array}{cc}
      \ssty \J_{\phi_{t-1}}^T & \\
      & \ssty \J_{\phi_t}^T
    \end{array}
  \right]
  \left[
    \begin{array}{cc}
      \ssty \I & \ssty -\I \\
      \ssty -\I & \ssty \I
    \end{array}
  \right]
  \left[
    \begin{array}{cc}
      \ssty \J_{\phi_{t-1}} & \\
      & \ssty \J_{\phi_t}
    \end{array}
  \right] \\
  &\approx
  \left[
    \begin{array}{cc}
      \ssty \A_{t-1} & \ssty -\C_{t-1}^T\C_t \\
      \ssty -\C_t^T\C_{t-1} & \ssty \A_t
    \end{array}
  \right],
\end{align}
where we've made the approximation $\U_{t-1}^T\U_t\approx\I$ for the off-diagonal 
terms, which is reasonable given 
the smoothness assumption on $\phi$.

Note that this procedure is only approximate because we linearize the maps in
the gradient computation and use the approximation $\U_t^T\U_{t-1}\approx\I$
for the Hessian calculation. It's usually better to exploit the actual
task map $\phi$ when available. But, as our experiments show below, 
this procedure can supply a good approximation when the task map isn't readily 
available.

\section{Experimental Results}

Our experiments first verify the theoretical convergence analysis of
Section~\ref{sec:TheoreticalConvergence} using controlled experiments on
kinematic objective terms under a realistic model of the manipulation system
pictured in Figure~\ref{fig:RealWorldResults}. We then empirically analyze the
optimization performance of both the new kinetic energy formulation of
Section~\ref{sec:ReformulatingKineticEnergy} and the corresponding performance
of the traditional energy formulation using the Cholesky approximation of
Section~\ref{sec:CholeskyApproximations}.  Finally, in
Section~\ref{sec:RealWorldExperiments}, we demonstrate the performance of our
full motion optimization system on the pictured Apollo manipulation platform
using both simulated and real-world executions. 

Our robotic platform is a dual arm manipulation system with two torque
controlled 7 DOF Kuka Lightweight arms with
4 DOF Barrett hands. Each experiment utilizes only the right side, and reduces
  the dimensionality of the configuration space to 8 DOF, either by restricting
  the finger motion to only the thumb/middle finger (locking the remaining
  fingers closed) or by using a single value to control all fingers
  simultaneously (locking the finger spread at a fixed maximal value).  Each
  real-world execution splines the $(\q, \qd, \qdd)$ trajectory using quintic
  splines and directly sends torque commands to the arm at a rate of 1kHz using
  inverse dynamics for active compliance.

\subsection{Controlled Hessian convergence experiment}

\begin{figure}[t]
  \centering
  \includegraphics[width=.495\columnwidth]{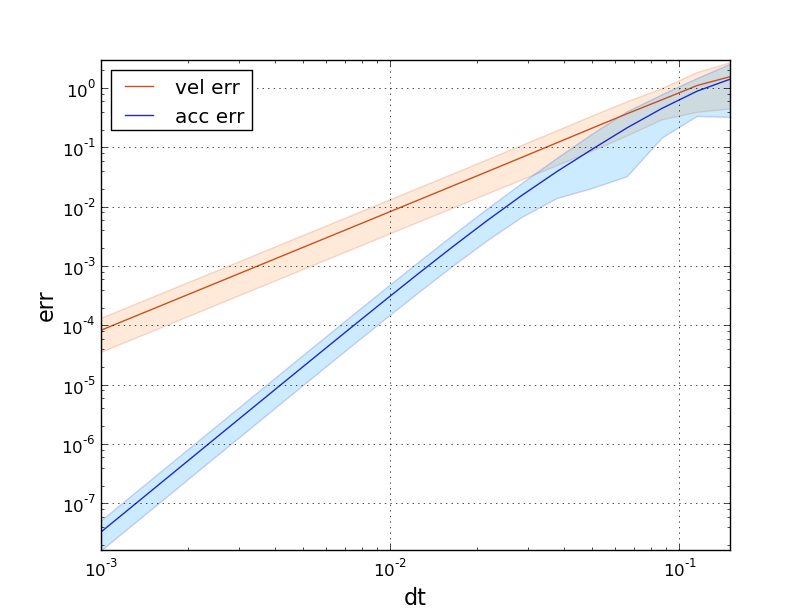}
  \includegraphics[width=.495\columnwidth]{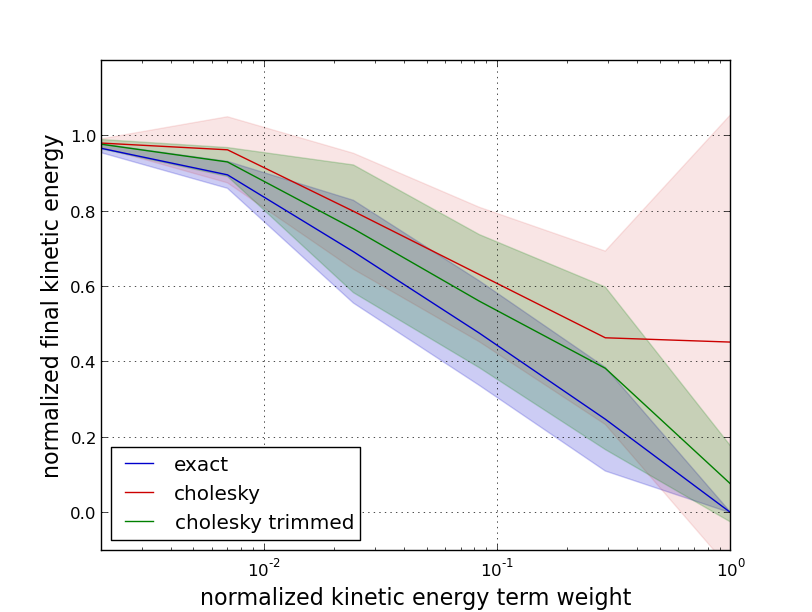}
  \vspace{-5mm}
  \caption{{\small {\bf Left:} Log-log plot
    showing convergence of full Hessians to Gauss-Newton Hessians
    for squared end-effector velocity and acceleration norm terms. 
    As predicted
    by Theorem~\ref{thm:MainTheorem}, the acceleration Hessian shows 
    a faster rate of convergence than the velocity Hessian
    as indicated by the relative slopes of the linear log-log relations.
    The plot shows 1 standard-deviation error bars for each progression.
    {\bf Right:}
    Plot showing the efficacy of \textit{exact} kinetic energy optimization
    using the Rigid Body Inertial Map derived in Section~\ref{sec:ReformulatingKineticEnergy},
    as well as the \textit{Cholesky} approximation presented in 
    Section~\ref{sec:CholeskyApproximations} (increasing the energy weight 
    decreases the final energy). The Cholesky approximation usually 
    performs well, but occasionally has difficulty. The \textit{trimmed} Cholesky 
    progression shows the behavior with outliers removed (see 
    the text for details). Both axes are normalized to lie between 0 and 1; 
    the maximum energy is chosen as the final trajectory energy resulting from optimizing without
    the energy term,
    and the minimum energy is the energy resulting from optimizing under
    the exact kinetic energy using 
    the largest term weight. The plot shows 1 standard-deviation 
    error bars for each progression.
    }}
  \vspace{-2mm}
  \label{fig:ControlledExperiments}
\end{figure}

Our first experiment directly addresses convergence and convergence rate
of the full Hessian to the Gauss-Newton Hessian as $\Delta t\rightarrow 0$.
We defined a smooth nonlinear $8$-dimensional configuration trajectory using the formula
\begin{align} \label{eqn:ExperimentalTrajectory}
  q^{(i)}(t) = \frac{\pi}{2} \sin\left(2\pi\sigma_i (t - {\scriptstyle\frac{1}{2}}) + \eta_i\right).
\end{align}
for each joint $i$ (ordered by distance from the base), 
where $\sigma_i\in\R$ ranges linearly from $.5$ to $2$ 
and $\eta_i\in\R$ ranges linearly from $0$ to $\pi$.
Time is in 
units of seconds.

This experiments analyzes the true Hessian of terms of the form
\begin{align}
  \spt{F}(\xi) = \sum_{t=1}^T \frac{1}{2}\|\x^{(k)}_t\|^2 dt,
\end{align}
where $\x = \phi_{\textrm{fk}}(\q)$ is the forward kinematics
map and $k$ denotes the $k^\textrm{th}$-order time derivative. 
For these 
experiments, we chose $k=1,2$ to examine task space 
velocities and accelerations. 

Figure~\ref{fig:ControlledExperiments}
shows the basic convergence of the $8\times 8$ diagonal blocks of 
the true Hessian (calculated using finite-differencing)
to the Gauss-Newton Hessian. The plots show the
\textit{normalized} error
\begin{align}
  \textrm{err}(\q_t) = 
      \frac{
        \|\mH_t - \wt{\mH}_t\|
       }{
        \|\mH_t\|
       },
\end{align}
where $\mH_t$ is the true Hessian block and $\wt{\mH}_t$ is the corresponding
Gauss-Newton Hessian block, 
both scaled by $\Delta t^{2k}$ as suggested by the theory. 
Here $\|\cdot\|$ denotes the Frobenius matrix norm.
We additionally verified separately that the scaled
full Hessian and Gauss-Newton Hessian both converge
individually 
to (the same) finite non-zero quantities as $\Delta \rightarrow 0$.

We evaluated the distribution of normalized Hessian block errors across 
the entire trajectory ($20$ linearly spaced choices of $t$ between $0$ and $1$) 
for each of the $20$ log-linearly spaced values of $\Delta t$ between $.15$ and $.001$.
The plots show the mean and one standard deviation error bars of the normalized
errors as a function of $\Delta t$ (in units of seconds).
Monomials appear linear in log-log scales; the linearity 
of these plots verifies the basic form of the predicted convergence rates
($O(\Delta t^2)$ for the velocity terms and $O(\Delta t^4)$ for the acceleration terms).
Moreover, the slope of the acceleration curve is twice that of the 
velocity curve verifying the differing relative rates of 
convergence as well.

\subsection{Controlled kinetic energy and Cholesky approximation experiment}

We implemented the inertial map derived in
Section~\ref{sec:ReformulatingKineticEnergy} using a slightly simplified model
of the mass distribution. We treat the upper arm and forearm as cylindrical
uniformly distributed rigid bodies of mass $9\:\textrm{kg}$ each and use
formulas for cylindrical rotational inertia around the appropriate axes to
calculate the requisite diagonal components of $\B$ as $b_1 =
.1875\:\textrm{kg\:m}^2$, and $b_2 = b_3 = .0324\:\textrm{kg\:m}^2$ along the
principle axes.  We model the inertial profile of the hand to be the same as
the upper and lower arms but with only $.9\:\textrm{kg}$ of mass.

For this experiment, each of $12$ trials started Apollo (simulated) in one of a
collection of initial configurations, and optimized a reach motion to touch one
of a collection of points in the 3D space under kinetic energy penalization
terms of varying strengths.  We used both the exact kinetic energy terms as
given by the task space velocity under the Rigid Body Inertial Map described in
Section~\ref{sec:ReformulatingKineticEnergy} and a Cholesky approximation of
the corresponding configuration space kinetic energy formulation using the
method described in Section~\ref{sec:CholeskyApproximations}. This problem is
intentionally simple to provide a more controlled experimental environment for
understanding the behavior of the kinetic energy term and it's Cholesky
approximation.  In addition to these energy terms, this model used small
squared norm terms on configuration space accelerations, initial velocities,
and terminal velocities, as well a postural term pulling back to a default
configuration for Null space resolution. 

The experiments used 6 separate weights for the kinetic energy term
log-linearly spaced between 1 and 500, in addition to a weight of 0 (removing
the term entirely). The plot shows these weights along the x-axis normalized
between 0 and 1.  For all nonzero weights, the exact energy term provided an
accurate Gauss-Newton Hessian estimate of the problem curvature, and,
therefore, optimized very well. We take its final cumulative kinetic energy
under the maximal weight as the minimum energy and the final kinetic energy
achieved with a weight of 0 as the maximum energy---the reported energy values
are scaled linearly between these minimum and maximum values to make them
comparable across all scenarios.  Figure~\ref{fig:ControlledExperiments} plots
the distribution of achieved normalized energy values as a function of term
weight (energy decreases with increased energy penalty term weight).

Generally, the Cholesky approximation performed well. For these trials, though,
we intentionally chose the start configuration and reaching target combinations
to stress the optimizer.  As a result, we found that the Cholesky approximation
had difficulty on $4$ of the $12$ trials especially as the term weight became
larger as can be seen in the figure. We, therefore, additionally plot the
performance of the distribution of the successful $8$ trials as ``cholesky
trimmed'' to show more generally the suboptimality of the approximation when
successful.

Understanding better when this Cholesky approximation is good is a subject of
future work. When it performs well, it converges very fast, in just slightly
more iterations (one or two) than the exact terms under Gauss-Newton Hessians.
Our empirical experience, though, emphasizes that, when explicit task maps can
be found, they are both fast and highly robust to problem variations.

\subsection{Experimental validation on a real-world system}
\label{sec:RealWorldExperiments}

\begin{figure*}[t]
  \centering
  \includegraphics[width=1.\textwidth]{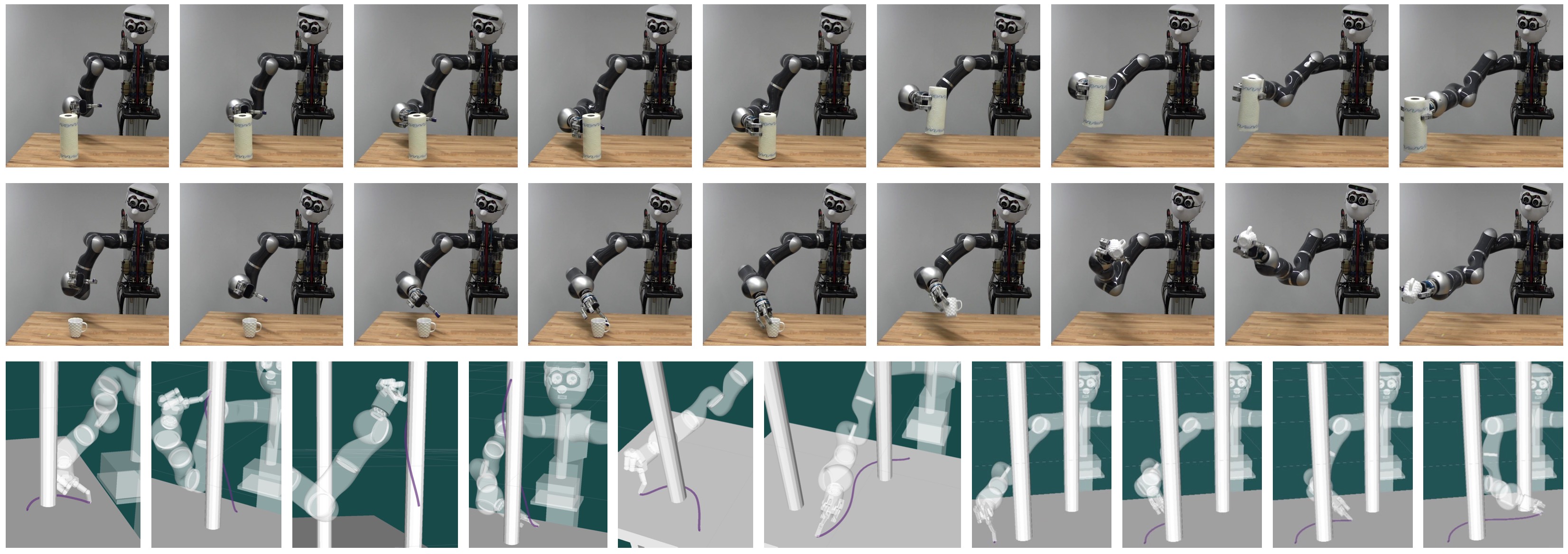}
  \vspace{-5mm}
  \caption{{\small 
    Row 1: Optimizing simultaneous reaching and cylindrical grasp preshaping,
    and a secondary lift motion with
    orientational constraints. Row 2: Optimizing reach and elliptical 
    preshaping for a more abnormally shaped object. Row 3: Optimizing
    motions around thin cylindrical objects using Gauss-Newton Hessians
    on geometric task maps
    mapping the workspace to a 9-dimensional latent Euclidean space
    modeling workspace geometry.
    The first 6 images 
    show the final configuration of the trajectory and the end-effector trail traced 
    out by the motion; the final 4 images show a sequence of 
    configurations for a single optimized trajectory moving among both
    cylinders.
    }}
  \vspace{-2mm}
  \label{fig:RealWorldResults}
\end{figure*}

Our final experimental demonstrations are on realistic problems using a full
implementation of the optimization system for optimized grasp reach and
preshaping and obstacle aware motions.  Figure~\ref{fig:RealWorldResults} shows
some of the resulting sequences.  Each optimization used $T=30$ and $\Delta t =
.1$ seconds.

Our motion model included the following objective terms and constraints:
{\small
\begin{itemize}
\item {\bf Configuration space derivatives penalties.} 
  $f(\qd, \qdd) = \alpha_1\|\qd\|^2 + \alpha_2\|\qdd\|^2$.
\item {\bf Task space velocity penalties.}  
  $f(\x, \xd) = \frac{1}{2}\|\frac{d}{dt}\phi(\x)\|^2$, where 
  $\phi:\R^3\rightarrow\R^n$ is a mapping to a
  workspace geometric space (see \cite{RIEMORatliff2015ICRA}). For the grasp preshape
  experiments, this geometric map was just the identity map.
\item {\bf Joint limit proximity penalties.} 
  $f_i(q_i) = 
  \left(\max\{0, 
              q_i - (q_\mathrm{max} - \epsilon), 
              (q_\mathrm{min} + \epsilon) - q_i\}\right)^2$,
  where $\epsilon > 0$ is a joint limit margin.
\item {\bf Posture potentials.} 
  $f(\q) = \frac{1}{2}\|\q - \q_{\mathrm{default}}\|^2$ pulling toward a 
  neutral default configuration
  $\q_{\mathrm{default}}$.
\item {\bf Kinetic energy.} These experiments use a simplified precursor
  to the more complete kinetic energy model defined in 
  Section~\ref{sec:ReformulatingKineticEnergy}. The robot was modeled
  as a collection of point masses at key locations along the robot's body.
\item {\bf Joint limit constraints.} We have explicit constraints on the joint 
  limits that prevent them from being violated in the final solution.
\item {\bf Obstacle constraints.} 
  All objects are modeled as analytical inequality constraints with 
  a margin. The end-effector, first knuckle, and second knuckle
  use margins of $0$cm, $1$cm, and $6$cm, respectively; the lower and upper 
  wrist joints use margins of $14$cm and $17$cm, respectively.
\item {\bf Goal constraint.} Reaching the 
  goal is enforced as a zero distance constraint on the goal proximity function.
  This strategy generalizes the goal set ideas of \cite{DraganGoalSetCHOMP2011}.
\end{itemize}
}

The optimizations to grasp preshape configurations used the weighted average
position of the fingertips and palm as the end-effector and chose the goal as a
point near the centroid of the object. A constraint surface surrounding the
object and reflecting the object's general shape enforced that the
fingertip positioning conform to a preshape configuration around the object
upon achieving the goal.
The cylindrical object used a cylindrical preshape surface in conjunction with
final configuration orientation constraint to align the hand to the principle
axis of the cylinder; the cup used a simple elliptical preshape surface without
orientation constraints.  Explicit quadratic potentials pulling the finger
joints toward open and close positions at times $\lfloor .75 T\rfloor$ and $T$, respectively,
implement the open and close behavior of the
hand as it approaches the object.  After achieving a successful preshape,
the robot simply squeezes using compliant control (implemented 
approximately here simply as
torque limits on the fingers and inverse-dynamics of the full arm) in the
manner of a pinch grasp. Once kinematic control over the object has been
achieved, the robot executes a lift and extend motion additionally optimized by
our motion optimizer. The cylindrical lift and extend motion used explicit
orientation constraints at each time step to keep the cylinder upright, while
the cup lift and extend motion did not. Both final motions used a uniform
upward potential in the workspace to implement the lift.  We used
Gauss-Newton Hessians wherever possible throughout this implementation.

The top two rows of Figure~\ref{fig:RealWorldResults} show examples of the
reach-preshape, grasp, and lift-extend motions. Each of these motions were
optimized starting from the zero motion trajectory in about .7
seconds.\footnote{All speed quotes here are for unoptimized code running on a
Linux virtual machine atop a 2012 Mac PowerBook---we designed the 
optimization and modeling
library for correctness and ease of use over speed.} This system
focused on the motion generation component and did not use a vision system for
object localization.  The results suggest that an optimization strategy
targeting viable preshapes that conform to the general contours of the object,
in conjunction with compliant grasping strategies, may constitute a successful
interaction approach for objects in unknown environments.
A more extensive empirical analysis of this approach integrating vision
is the subject of future work.

The final row of the figure shows obstacle avoidance behaviors optimized using
latent Euclidean task maps mapping the workspace to a higher-dimensional space
that encodes how environmental obstacles warp the geometry of the workspace
(see \cite{RIEMORatliff2015ICRA}). Each of these motions took between $.3$ and
$1$ seconds of computation, using Gauss-Newton Hessians wherever possible.

%%%%%%%%%%%%%%%%%%%%%%%%%%%%%%%%%%%%%%%%%%%%%%%%%%%%%%%%%%%%%%%%%%%%%%%%%%%%%%%%

\section*{ACKNOWLEDGMENT}

{\small 
The experimental work on the physical Apollo system in this paper wouldn't have
been possible without the efforts of Ludovic Righetti and Mrinal Kalakrishnan,
who implemented much of the underlying control architecture used for this project.
Nathan Ratliff was jointly affiliated with the Max Planck and U. Stuttgart during 
the execution of much of this work.
}

{\small
\bibliography{refs}

\begin{thebibliography}{10}

\bibitem{DraganGoalSetCHOMP2011}
Anca Dragan, Nathan Ratliff, and Siddhartha Srinivasa.
\newblock Manipulation planning with goal sets using constrained trajectory
  optimization.
\newblock In {\em 2011 IEEE International Conference on Robotics and
  Automation}, May 2011.

\bibitem{DRCIntegratedSystemTodorov2013}
Tom Erez, Kendall Lowrey, Yuval Tassa, Vikash Kumar, Svetoslav Kolev, and
  Emanuel Todorov.
\newblock An integrated system for real-time model-predictive control of
  humanoid robots.
\newblock In {\em IEEE/RAS International Conference on Humanoid Robots}, 2013.

\bibitem{ClosedFormDynamicsDerivativesOtt2013}
Gianluca Garofalo, Christian Ott, and Alin Albu-Sch{\"a}ffer.
\newblock On the closed form computation of the dynamic matrices and their
  differentiations.
\newblock In {\em IEEE/RSJ International Conference on Intelligent Robots and
  Systems (IROS)}, November 2013.

\bibitem{Kalakrishnan_RAIIC_2011}
M.~Kalakrishnan, S.~Chitta, E.~Theodorou, P.~Pastor, and S.~Schaal.
\newblock {STOMP}: Stochastic trajectory optimization for motion planning.
\newblock In {\em Robotics and Automation (ICRA), 2011 IEEE International
  Conference on}, 2011.

\bibitem{LeeSmoothManifolds2012}
John~M. Lee.
\newblock {\em Introduction to Smooth Manifolds}.
\newblock Springer, 2nd edition, 2002.

\bibitem{NashEmbeddingTheorem1956}
John Nash.
\newblock The imbedding problem for {R}iemannian manifolds.
\newblock {\em Ann. Math.}, 63:20--63, 1956.

\bibitem{NocedalWright2006}
Jorge Nocedal and Stephen Wright.
\newblock {\em Numerical Optimization}.
\newblock Springer, 2006.

\bibitem{RIEMORatliff2015}
Nathan Ratliff, Marc Toussaint, and Stefan Schaal.
\newblock Riemannian motion optimization.
\newblock Technical report, Max Planck Institute for Intelligent Systems, 2015.

\bibitem{RIEMORatliff2015ICRA}
Nathan Ratliff, Marc Toussaint, and Stefan Schaal.
\newblock Understanding the geometry of workspace obstacles in motion
  optimization.
\newblock In {\em IEEE International Conference on Robotics and Automation},
  2015.

\bibitem{RatliffCHOMP2009}
Nathan Ratliff, Matthew Zucker, J.~Andrew~(Drew) Bagnell, and Siddhartha
  Srinivasa.
\newblock {CHOMP}: Gradient optimization techniques for efficient motion
  planning.
\newblock In {\em IEEE International Conference on Robotics and Automation
  (ICRA)}, May 2009.

\bibitem{AbbeelTrajopt2013}
John~D. Schulman, Jonathan Ho, Alex Lee, Ibrahim Awwal, Henry Bradlow, and
  Pieter Abbeel.
\newblock Finding locally optimal, collision-free trajectories with sequential
  convex optimization.
\newblock In {\em In the proceedings of Robotics: Science and Systems (RSS)},
  2013.

\bibitem{KinematicsDynamicsControl10}
Bruno Siciliano, Lorenzo Sciavicco, Luigi Villani, and Giuseppe Oriolo.
\newblock {\em Robotics: Modelling, Planning and Control}.
\newblock Springer, second edition, 2010.

\bibitem{iLQGTodorov05}
E.~Todorov and W.~Li.
\newblock A generalized iterative lqg method for locally-optimal feedback
  control of constrained nonlinear stochastic systems.
\newblock In {\em In proceedings of the American Control Conference}, volume~1,
  pages 300--306, 2005.

\bibitem{ToussaintTrajOptICML2009}
Marc Toussaint.
\newblock Robot trajectory optimization using approximate inference.
\newblock In {\em (ICML 2009)}, pages 1049--1056. ACM, 2009.

\bibitem{KOMOToussaint2014}
Marc Toussaint.
\newblock Newton methods for k-order {M}arkov constrained motion problems.
\newblock {\em CoRR}, abs/1407.0414, 2014.

\end{thebibliography}
\bibliographystyle{plain}
}

%%%%%%%%%%%%%%%%%%%%%%%%%%%%%%%%%%%%%%%%%%%%%%%%%%%%%%%%%%%%%%%%%%%%%%%%%%%%%%%%
\section*{APPENDIX} 
\setcounter{section}{1}
\setcounter{subsection}{0}
{
\subsection{Exploiting Rigid Body Structure} \label{sec:ExploitingRigidBodyStructure}

Here we provide some details on the decomposition leading 
to Equation~\ref{eqn:RigidBodyEnergyDecomposition}.
First, we note that by the definition of center-of-mass,
\begin{align*}
  \x_c 
  &= \frac{1}{M} \int \x(\mat{u}) \rho(\mat{u}) d\mat{u}
  = \frac{1}{M} \int \left(\x_c + \sum_i u_i \e_i\right) \rho(\mat{u}) d\mat{u} \\
  &= \x_c + \frac{1}{M}\sum_i \left(\int u_i \rho(\mat{u}) d\mat{u}\right) \e_i.
\end{align*}
And since the set $\spt{B} = \{\e_i\}_{i=1}^3$ forms an orthonormal basis, it must be that
\begin{align} \label{eqn:COMVanish}
  \int u_i \rho(\mat{u}) d\mat{u} = 0\ \ \ \textrm{for all}\ \ i=1,2,3.
\end{align}
Now, expanding the kinetic energy expression, we get
\begin{align*}
  &\frac{1}{2}\int \rho(\mat{u})\:\|\xd\|^2d\mat{u}
  = \frac{1}{2}\int \rho(\mat{u})\:\|\xd_c + \sum_i u_i\ed_i\|^2 d\mat{u} \\
  &\ \ = \frac{1}{2}\int\rho(\mat{u})\left(\|\xd_c\|^2 + 2\sum_i u_i\xd_c^T\ed_i + \|\sum_i u_i\ed_i\|^2\right)\\
  &\ \ = \frac{1}{2} M\|\xd_c\|^2
    + \sum_i\xd_c^T\ed_i\int u_i\rho(\mat{u})\:d\mat{u}\\
  &\ \ \ \ \ \ \ \ \ \ \ \ \ \ \ \ \ \ \ \ \ \ \ \ \ \ 
  + \frac{1}{2}\sum_{ij} \left(\int u_iu_j\:\rho(\mat{u}) d\mat{u}\right) \ed_i^T\ed_j
\end{align*}
The second term vanishes because of Equation~\ref{eqn:COMVanish}, and the first and last terms,
together, constitute Equation~\ref{eqn:RigidBodyEnergyDecomposition}.

\subsection{Relationship between the Distributional Inertia Matrix and the traditional
  Inertia Matrix} \label{sec:RelationshipBetweenInertiaMatrices}

It's fairly straightforward to show, simply from the definition of the traditional 
inertial matrix\footnote{Not to be confused notationally with the identity 
matrix---it's common in physics and robotics literature to denote 
the rigid body inertia matrix as $\I$, so we follow that convention here.} 
$\I$ \cite{KinematicsDynamicsControl10}, that the entries of the Distributional Inertia
Matrix defined in Section~\ref{sec:ReformulatingKineticEnergy} are given by
\begin{align*}
  \B = \left[
  { \tiny
    \begin{array}{ccc}
      \frac{1}{2}(I_{22} + I_{33} - I_{11}) & -I_{12} & -I_{13} \\
      -I_{21} & \frac{1}{2}(I_{11} + I_{33} - I_{22}) & -I_{23} \\
      -I_{31} & -I_{32} & \frac{1}{2}(I_{11} + I_{22} - I_{33})
    \end{array}
  }
  \right].
\end{align*}
The off-diagonal entries of $\B$ are just the negatives of the off-diagonal
entries of $\I$, so $\B$ is diagonal if and only if $\I$ is diagonal.
}

\end{document}